\documentclass[letterpaper]{article} 
\usepackage{aaai24}  
\usepackage{times}  
\usepackage{helvet}  
\usepackage{courier}  
\usepackage[hyphens]{url}  
\usepackage{graphicx} 
\usepackage{natbib}  
\usepackage{caption} 
\usepackage{algorithm}
\usepackage{algorithmic}
\usepackage{multirow}
\usepackage{multicol}
\usepackage{makecell}
\usepackage{booktabs}

\usepackage{newfloat}
\usepackage{listings}
\DeclareCaptionStyle{ruled}{labelfont=normalfont,labelsep=colon,strut=off} 
\floatstyle{ruled}
\newfloat{listing}{tb}{lst}{}
\floatname{listing}{Listing}
\title{Building Variable-sized Models via Learngene Pool}
\author {
    Boyu Shi,
    Shiyu Xia,
    Xu Yang\thanks{Co corresponding author.},
    Haokun Chen,
    Zhiqiang Kou,
    Xin Geng\thanks{Co corresponding author.}
}
\affiliations {
    School of Computer Science and Engineering,
    Southeast University, Nanjing 210096, China\\
    Key Laboratory of New Generation Artificial Intelligence Technology and Its Interdisciplinary Applications (Southeast University), Ministry of Education, China\\
    \{shiboyu, shiyu\_xia, 101013120, chenhaokun, zhiqiang\_kou, xgeng\}@seu.edu.cn
}

\usepackage{bibentry}

\begin{document}

\maketitle

\begin{abstract}
Recently, Stitchable Neural Networks (SN-Net) is proposed to stitch some pre-trained networks for quickly building numerous networks with different complexity and performance trade-offs. In this way, the burdens of designing or training the variable-sized networks, which can be used in application scenarios with diverse resource constraints, are alleviated. However, SN-Net still faces a few challenges. 1) Stitching from multiple independently pre-trained anchors introduces high storage resource consumption. 2) SN-Net faces challenges to build smaller models for low resource constraints. 3). SN-Net uses an unlearned initialization method for stitch layers, limiting the final performance.
To overcome these challenges, motivated by the recently proposed Learngene framework, we propose a novel method called \textbf{Learngene Pool}. Briefly, Learngene distills the critical knowledge from a large pre-trained model into a small part (termed as \textbf{learngene}) and then expands this small part into a few variable-sized models. In our proposed method, we distill one pre-trained large model into multiple small models whose network blocks are used as \textbf{learngene instances} to construct the learngene pool. Since only one large model is used, we do not need to store more large models as SN-Net and after distilling, smaller learngene instances can be created to build small models to satisfy low resource constraints. We also insert learnable transformation matrices between the instances to stitch them into variable-sized models to improve the performance of these models. Exhaustive experiments have been implemented and the results validate the effectiveness of the proposed \textbf{Learngene Pool} compared with SN-Net.
\end{abstract}

\section{Introduction}
\begin{figure*}[t]
	\centering
	\includegraphics[width=0.8\textwidth]{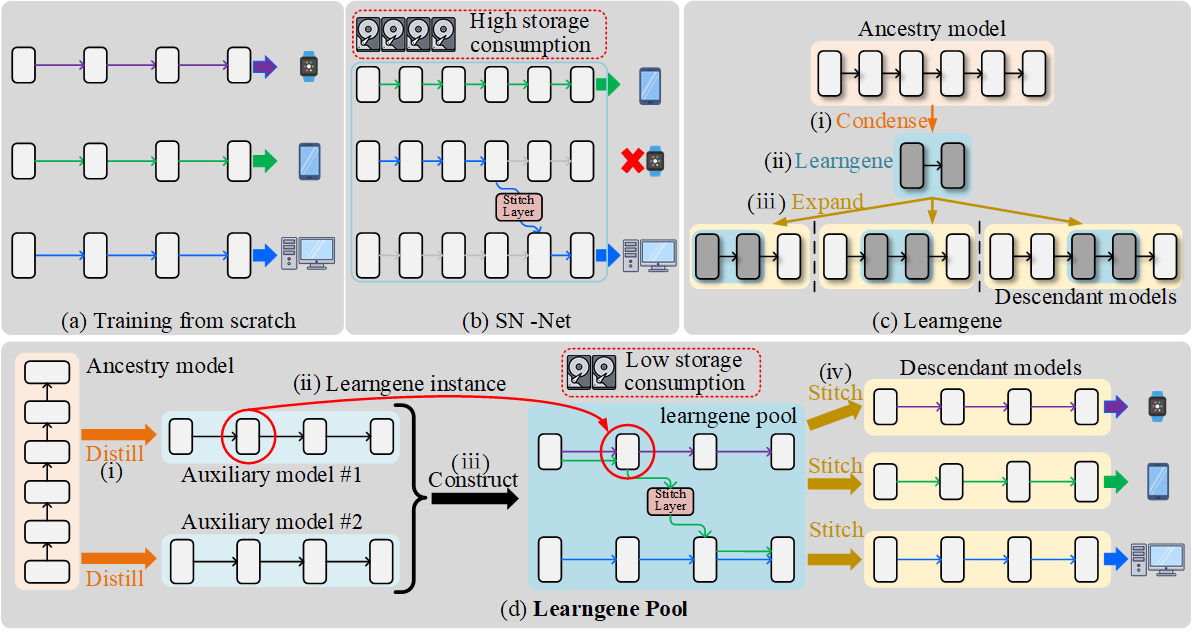}
	\caption[Figure]{(a) Designing the specific model and training from scratch for all resource constraints consumes lots of resources. (b) SN-Net cannot adapt to low resource constraints and consumes lots of storage resources to save all anchors. (c) The framework of Learngene. (\romannumeral1 \& \romannumeral2) Learngene condenses critical parts, termed learngene, from the large ancestry model. (\romannumeral3) The extracted learngene is expanded into small or medium-sized descendant models. (d) The overall process of \textbf{Learngene Pool}. (\romannumeral1) Distilling one ancestry into multiple smaller auxiliary models. (\romannumeral2 \& \romannumeral3) Selecting each individual block in the auxiliary models as an instance to construct the learngene pool. (\romannumeral4) The variable-sized descendant models are built by stitching from the learngene pool for various resource constraints.}
	\label{fig:motivation}
\end{figure*}

Deep learning models \cite{lecun2015deep,Dehghani2023ScalingVT} have demonstrated their applicability and significance in various fields, being deployed on diverse devices like watches, smartphones, PCs, etc \cite{Gholami2018SqueezeNextHN,Howard2017MobileNetsEC}. However, the diverse resource constraints of these devices lead to variations in the scale of deep learning models \cite{simonyan2014very,he2016deep,dosovitskiy2020image}. 
To design models with different scales, conventional deep learning approaches typically involve manual crafting of specific model sizes for each resource constraint  \cite{Li2022EfficientFormerVT,Li2022RethinkingVT,Mehta2021MobileViTLG,Li2020MicroNetTI}, necessitating training from scratch (Figure \ref{fig:motivation} (a)) or compressing the huge models into smaller models \cite{Fang2022UpT1,Hameed2021ConvolutionalNN,Zhao2022DecoupledKD,Fang2021MosaickingTD,Frantar2022OptimalBC,Zhang2022AdvancingMP}. 
However, these approaches are time-consuming and impractical for generating differently scaled models

To address this issue, a novel method called Stitchable Neural Network (SN-Net) \cite{pan2023stitchable} has been proposed. 
Unlike the conventional approach of training individual scale-specific models, SN-Net \cite{pan2023stitchable} leverages pretrained models to construct variable-sized models, resulting in significant time savings. 
Specifically, SN-Net \cite{pan2023stitchable} selects pretrained varying-size models, referred to as anchors, from a model family (e.g., DeiT-Ti/S/B \cite{touvron2021training}), and performs stitching operations among these anchors (Figure \ref{fig:motivation} (b)). The stitching is accomplished by introducing a $1 \times 1$ convolution layer, termed a stitch layer, to establish a new forward propagation path between the two adjacent anchors. 
By stitching varied-sized blocks of various anchors, SN-Net \cite{pan2023stitchable} can generate models of varying sizes.

However, SN-Net \cite{pan2023stitchable} still possesses several limitations. 
Firstly, SN-Net \cite{pan2023stitchable} performs stitching operations among two or more independently-trained anchors (Figure \ref{fig:motivation} (b)), resulting in substantial consumption of storage resources.
Additionally, the minimum scale of the model stitched by SN-Net \cite{pan2023stitchable} depends on the selected smallest anchor. If the parameters of the smallest anchor are not sufficiently small, SN-Net \cite{pan2023stitchable} is unable to generate smaller models. 
For example, when stitching between DeiT-Ti/S/B \cite{touvron2021training}, the size of the stitched model is greater than or equal to that of DeiT-Tiny (5.7M) \cite{touvron2021training}. Given a smartwatch with a target resource constraint under 5M, SN-Net cannot build a model to satisfy this constraint, as shown in figure \ref{fig:motivation} (b), 
Finally, SN-Net computes the parameters of stitch layers by employing the least squares method from the feature maps of the anchors' outputs: $\min \|F_{I_j}(x) {W}-{F_{I_{j+1}}}(x)\|$, where $F_{I_j}(x)$ and $F_{I_{j+1}}(x)$ are the output feature maps of two anchors, $W$ means the transformation matrix and $x$ is the input samples. However, small anchors are hard to output the proper intermediate feature map, which makes this unlearnable initialization method limit the performance of the stitching layers.

One recently proposed method called Learngene \cite{wang2022learngene} shows promise in mitigating the limitations of SN-Net \cite{pan2023stitchable}. Figure \ref{fig:motivation} (c) illustrates the overall process of the Learngene. In Figure \ref{fig:motivation} (c)(\romannumeral1), Learngene first condenses a larger, well-trained model, termed ancestry model, into a tiny critical part known as learngene (Figure \ref{fig:motivation} (c)(\romannumeral2)), which contains essential information from the ancestry model. 
Since the learngene is a tiny part, it consumes few storage resources. 
Subsequently, learngene is expanded to create many variable-sized models for downstream tasks, which are called descendant models, as shown in Figure \ref{fig:motivation} (c)(\romannumeral3). 
Similar to SN-Net \cite{pan2023stitchable}, Learngene utilizes well-trained models to build variable-sized models without training from scratch. 
Differently, Learngene generates models of different sizes by expanding the critical component, i.e., learngene, enabling the constructed models to cover smaller sizes. 
This effectively addresses the limitation of SN-Net \cite{pan2023stitchable}, which faces challenges to build smaller models for low resource constraints.

However, the vanilla Learngene \cite{wang2022learngene} takes a simplistic approach by extracting the last three layers of the ancestry model as learngene and combining them with randomly initialized layers to build descendant models. This approach is inadequate to fully address the challenges encountered by SN-Net \cite{pan2023stitchable}.
Inspired by the principles of Learngene, we propose a novel approach called \textbf{Learngene Pool} to comprehensively overcome the limitations of SN-Net \cite{pan2023stitchable}.

\textbf{Learngene Pool} enables the construction of variable-sized models from the learngene pool, and the overall process is shown in Figure \ref{fig:motivation} (d). To establish a learngene pool, we begin by selecting a well-trained ancestry model. 
In this study, we adopt DeiT-Base \cite{touvron2021training} as the ancestry model. 
Then, we design multiple models, referred to as the `auxiliary models', to condense the critical knowledge of the ancestry model into smaller learngene in two ways: reducing the number of blocks and lowering the output dimensions of the blocks. In Figure \ref{fig:motivation} (d)(\romannumeral1), the learngene is extracted by distilling from the ancestry model to the auxiliary models. During distillation, multiple learnable transformation matrices are designed to match the output dimensions between the ancestry model and the auxiliary models. 

After training, in Figure \ref{fig:motivation} (d)(\romannumeral2), each block of the auxiliary models is selected as `learngene instances' (abbreviated as the `instance' for convenience). 
Subsequently, in Figure \ref{fig:motivation} (d)(\romannumeral3), these selected instances collectively construct the learngene pool. Instances in it are arranged in the order of the output dimensions.
Then, as shown in Figure \ref{fig:motivation} (d)(\romannumeral4), we stitch learngene instances from the learngene pool to generate descendant models that meet various resource constraints. Similar to SN-Net \cite{pan2023stitchable}, we also insert stitch layers between different learngene instances to match their output dimensions. However, we initialize the stitch layers using the parameters of the transformation matrices learned during the distillation process.

Empirically, compared to SN-Net \cite{pan2023stitchable}, \textbf{Learngene Pool} employs a reduced number of instances, thus consuming fewer storage resources. 
Specifically, the learngenen pool which contains 12 and 18 instances respectively save around 59.6\% and 40.1\% storage resources compared to SN-Net. 
Furthermore, fewer instances in the learngene pool facilitate the construction of descendant models with fewer parameters. 
For instance, while the DeiT-based SN-Net only builds the models exceeding 5.7M parameters, the 12 and 18 instances learngenen pool can construct smaller descendant models: 3.05M and 4.38M parameters, respectively.
Moreover, we initialize the stitch layers in the learngene pool by the learned block-based transformation matrices, resulting in improved final performance.
Additionally, when compare \textbf{Learngene Pool} and SN-Net at the same storage resource costs, the 12 instances learngene pool improves the SN-Net results from 67.89\% to 75.05\% at 44.04M parameters, and the 18 instances learngene pool enhances results from 69.11\% to 77.42\% at 65.03M parameters.


\begin{figure*}[t]
	\centering
	\includegraphics[width=0.9\textwidth]{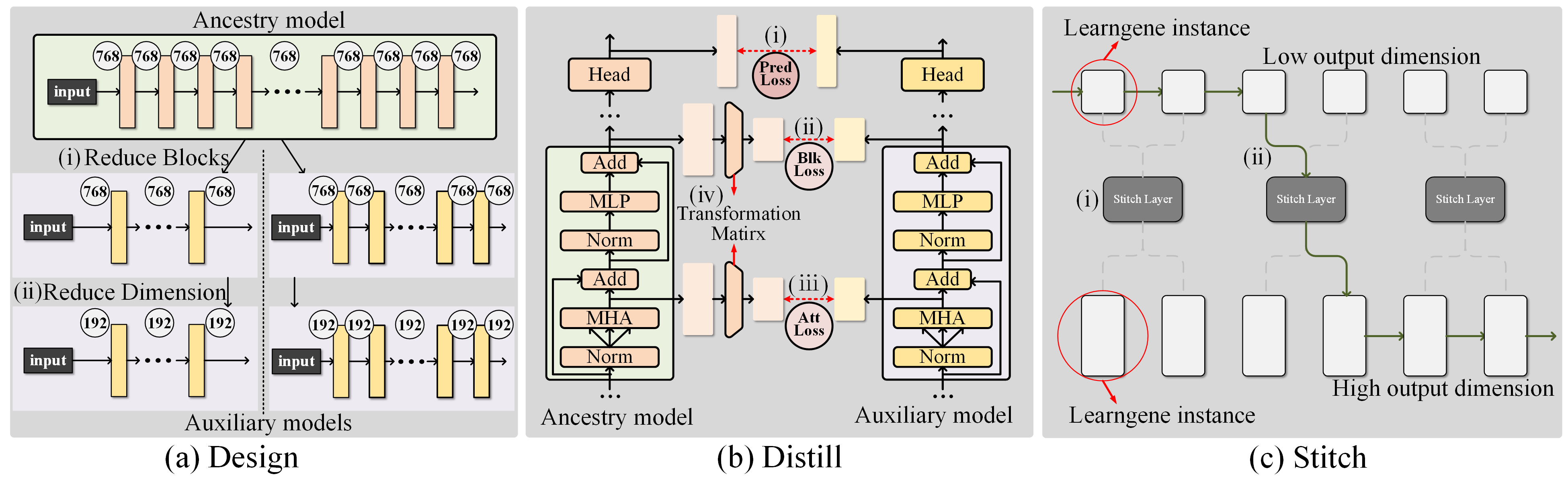}
	\caption[Figure]{The technical details of the \textbf{Learngene Pool}. (a) The designing way of the auxiliary models. (\romannumeral1) Reducing the number of blocks. (\romannumeral2) Reducing the output dimensions based on the ancestry model. Two groups of auxiliary models with various numbers of blocks are designed. (b) Distillation ways. (\romannumeral1 \& \romannumeral2 \& \romannumeral3) Measuring the outputs of the head operation, block, and self-attention operation between the ancestry model and the auxiliary models. (\romannumeral4) Adopting the transformation matrices to match both output dimensions. (c) Stitching ways. (\romannumeral1) The descendant model is constructed by sequentially stitching instances in the learngene pool. (\romannumeral2) Stitching from the smaller instances to the larger instances.}
	\label{fig:overall}
\end{figure*}

\section{Related Work}
\subsection{Learngene}
The vanilla Learngene approach \cite{wang2022learngene} extracts layers with stable gradients during the training of the ancestry model as learngene. Since higher-level semantic layers of the ancestry model have stable gradients, they are extracted as learngene. Then, the extracted learngene layers are combined with randomly initialized other layers to build the descendant model for the downstream tasks. 

Recently, a new Learngene method \cite{Wang2023LearngeneIC} has been proposed, which is based on the observation \cite{Selvaraju_Cogswell_Das_Vedantam_Parikh_Batra_2020,Jiang_Zhang_Hou_Cheng_Wei_2021} that integral layers contain critical knowledge.
To extract learngene, this work first designs the pseudo descendant model for the ancestry model and trains them with the same task. Then, a meta-network is introduced to calculate the layer similarity score between the two models by following the meta-learning mechanism \cite{Vanschoren_2018,Finn_Abbeel_Levine_2017}. The layers which have high similarity scores in the ancestry model are extracted as learngene layers. The extracted learngene layers are then stacked with various randomly initialized layers to build the descendant models.

\subsection{Model Stitching}
The concept of model stitching \cite{Lenc2014UnderstandingIR} is introduced to build the new model by connecting the initial layers of one trained network with the final layers of another trained network with stitching layers. It aims to explore similarities in internal representations across different neural networks.
Sequentially, \cite{Bansal2021RevisitingMS,Csiszrik2021SimilarityAM} applies model stitching to build new networks to further study the network representations.
A recent work called \cite{Yang2022DeepMR} is proposed using model stitching to create a customized network for specific downstream tasks by dissecting and reassembling well-trained models.

Unlike previous approaches, SN-Net \cite{pan2023stitchable} is proposed to build variable-sized models by stitching from the pretrained models (termed anchors). The stitch process is achieved by inserting $1 \times 1$ convolution layers, referring to stitch layers, into these anchors. Since anchors consist of blocks of varying sizes, stitching these blocks can build various models of diverse sizes.

However, saving multiple anchors consumes lots of storage resources. 
Moreover, large anchors limit the minimum scale of the generated models, thus restricting their adaptability to low resource constraints.
The stitch layers in SN-Net are initialized in unlearnable ways, which decreases the final performance of the built models.

\section{Methodology}
To alleviate the limitations of SN-Net, we propose a novel approach called \textbf{Learngene Pool}. This method can be divided into two main procedures: constructing the learngene pool and building the descendant model. The technical details are depicted in Figure \ref{fig:overall}.
In this section, we first provide the detailed construction process of the learngene pool. Then, we illustrate the procedure for building the variable-sized descendant models.

\subsection{Constructing the Learngene Pool}
\textbf{Selecting the Ancestry Model.}  To construct the learngene pool, we first carefully choose a suitable ancestry model. In our study, we adopt DeiT-Base \cite{touvron2021training} as the ancestry model for the following reason. DeiT-Base has more parameters in the DeiT family \cite{touvron2021training}, thus granting it to learn superior representations on pretrained tasks such as ImageNet \cite{russakovsky2015imagenet}. This advantage enables us to extract more effective learngenes from it, which is crucial in constructing the learngene pool. The ancestry model $F_{anc}^L$ with $L=12$ blocks can be denoted as:
\begin{equation}
F_{anc}^{12} = f_{anc}^H \circ f_{anc}^{12} \circ \cdots f_{anc}^i \cdots \circ f_{anc}^1 \circ f_{anc}^{PE},
\end{equation}

\textbf{Design Auxiliary Models.} To carry the critical knowledge (i.e., learngenes) extracted from the ancestry model, we design two auxiliary models for each learngene pool: 
the first one is to reduce the number of blocks of the ancestry model (Figure \ref{fig:overall} (a)(\romannumeral1)), and the second one is to further decrease the output dimensions of each block based on the first one to obtain a smaller auxiliary model (Figure \ref{fig:overall} (a)(\romannumeral2)). 
For example, the output dimensions are reduced from 768 to 192. 
We denote the auxiliary model with $l$ blocks as:
\begin{equation}
F_{aux}^{l} = f_{aux}^H \circ f_{aux}^{l} \circ \cdots f_{aux}^i \cdots \circ f_{aux}^1 \circ f_a^{PE},
\end{equation}
where $f_{aux}^i$ represents the $i$-th block of the auxiliary model. 
In this study, we conduct two experiments by designing two groups of auxiliary models (Figure \ref{fig:overall} (a)) for validating the effectiveness of \textbf{Learngene Pool}. 
The first group designs auxiliary models with 6 blocks, and the second group designs auxiliary models with 9 blocks.

\textbf{Training the Auxiliary Models.} After designing, we extract the learngene from the ancestry model to the auxiliary model by distillation. To this end, we adopt three types of distillation loss functions to transfer learngene, which is inspired by TinyBERT \cite{Jiao2019TinyBERTDB}. 

Firstly, we utilize prediction-layer-based distillation \cite{Hinton2015DistillingTK} to extract critical information in the prediction layer from the ancestry model into the auxiliary models. This process is achieved by adopting the soft cross-entropy loss $\mathrm{CE_{soft}}(\cdot)$ between the ancestry model's output logits $P_{anc}$ against the auxiliary models' output logits $P_{aux}$, and the objective is defined as:
\begin{equation}
\label{qe:pred}
\mathcal{L}_{pred} = \mathrm{CE_{soft}}\left(P_{anc} / \tau, P_{aux} / \tau\right), 
\end{equation}
where $\tau$ is the temperature value of the distillation. In this study, $\tau$ has the same value as TinyBERT \cite{Jiao2019TinyBERTDB}, which is equal to 1. This process is illustrated in \ref{fig:overall} (b)(\romannumeral1).

Secondly, we employ block-based distillation, aimed at transferring key knowledge contained in the blocks from the ancestry model to the blocks of the auxiliary models. This process is formulated as:
\begin{equation}
\mathcal{L}_{blk} = \mathrm{MSE}\left(B_{anc}W, B_{aux}\right), 
\end{equation}
where $B_{anc}$ and $B_{aux}$ refer to the blocks' output of the ancestry model and the auxiliary models respectively, as shown in Figure \ref{fig:overall} (b)(\romannumeral2). The matrix $W\in R^{d \times d^{\prime}}$ is a learnable block-based transformation matrix, which transforms the output dimension $d$ of $B_{anc}$ to match the output dimension $d^{\prime}$ of $B_{aux}$ (Figure \ref{fig:overall} (b)(\romannumeral4)).

Finally, we employ attention-based distillation, which encourages the auxiliary models to learn the informative representations of input data captured by the attention layers \cite{dosovitskiy2020image} of the ancestry model. Specifically, as depicted in Figure \ref{fig:overall} (b)(\romannumeral3), the output of the multi-head attention layer in the auxiliary models $A_{aux}$ is aligned with the output of the corresponding multi-head attention layer in the ancestry model $A_{anc}$. This function is denoted as:
\begin{equation}
\mathcal{L}_{att} = \mathrm{MSE}\left(A_{anc}M, A_{aux}\right), 
\end{equation}
where $M \in R ^{d \times d^{\prime}}$ is an attention-based transformation matrix, which transforms the output dimension $d$ of $A_{anc}$ into the same dimension $d^{\prime}$ as $A_{aux}$ (Figure \ref{fig:overall} (b)(\romannumeral4)). Note that, when $d = d^{\prime}$, both $M$ and $W$ are the simple identity matrices. Therefore, the total distillation loss is:
\begin{equation}
\mathcal{L}_{dis} = \mathcal{L}_{att} + \mathcal{L}_{blk} + \mathcal{L}_{pred}.
\end{equation}

In addition to distillation, we also pre-train the auxiliary models:
\begin{equation}
\label{eq:cls}
\mathcal{L}_{cls} = \mathrm{CE}\left(y_c, F_{aux}(x)\right), 
\end{equation}
where $x$ represents the input data and $y_c$ denotes the label belonging to category $c$. Then, the total loss function of training the auxiliary models is:
\begin{equation}
\label{eq:loss}
\mathcal{L} = \alpha \mathcal{L}_{cls} + (1-\alpha)\mathcal{L}_{dis}.
\end{equation}

\textbf{Dense Distillation.} 
We default to adopt the last block to achieve attention-based and block-based distillations.
However, since the auxiliary models which output lower dimensions than the ancestry model have poor learning abilities, distilling the final block is insufficient for them to fully extract whole critical knowledge from the large ancestry model.
To address this challenge, we introduce a dense distillation, which means taking multiple blocks to calculate the distillation loss functions.
Specifically, we categorize both the ancestry model and auxiliary models into three levels: the low level, the middle level, and the high level based on the observation that different layers within the model exhibit varying abilities \cite{Zeiler2013VisualizingAU,Zhang2019AreAL}. 
At each level, we distill information from the final block of the ancestry model to the corresponding final block in the auxiliary models. 
For example, in the case of an auxiliary model with 6 blocks, we distill the 4th, 8th, and 12th blocks of the ancestry model into the 2nd, 4th, and 6th blocks of the auxiliary models. In this way, the whole critical knowledge of the ancestry model can be extracted into the auxiliary models with lower output dimensions.

After training, the critical knowledge of the ancestry model (i.e., learngene) has been extracted into all blocks of the auxiliary models. 
Therefore, for each auxiliary model, we select each block as one learngene instance to build the learngene pool. 
Within the learngene pool, learngene instances from the same auxiliary model are on a single line, arranged in order of their position in the auxiliary model. Also, the output dimension of learngene instances increases row by row in the learngen pool. 
Additionally, as shown in Figure \ref{fig:overall} (c)(\romannumeral1), we insert multiple stitch layers between different rows in the learngene pool to transform outputs from one learngene instance to another. 

\subsection{Building the Descendant Models}
\textbf{Initialization of the Stitch Layers.} 
SN-Net \cite{pan2023stitchable} calculates the parameters by the least squares method to initialize the stitch layers. However, this cannot work well when the anchor is small. To enhance the performance of the stitch layers in the learngene pool, we initialize stitch layers by the parameters of block-based transformation matrices obtained from the distillation process. 
Specifically, for the auxiliary models with lower output dimensions, we introduce 3 block-based transformation matrices for distilling. We average them to obtain $W \in R^{192 \times 768}$ and employ it to initialize all stitch layers between learngene instances with 768 dimensions and those with 192 dimensions.

\textbf{Finetuning the Learngene Pool.}
We conduct additional training of the learngene pool to enhance its performance. The training progress takes inspiration from the work \cite{Guo2019SinglePO}, where we randomly sample a single stitching path from the learngene pool and execute a single backward propagation step each time. This iterative process continues until the training reaches the end.
To further improve the performance of the learngene pool, we also employ the pretrained DeiT-Base \cite{touvron2021training} to guide the training of the learngene pool.
The pipeline of building and training the learngene pool is summarized in Alg.\ref{alg:learngene_pool}.
\begin{algorithm}[t]
    \renewcommand{\algorithmicrequire}{\textbf{Input:}}
    \renewcommand{\algorithmicensure}{\textbf{Output:}}
    \caption{Building and Finetuning the Learngene Pool}
    \label{alg:learngene_pool}
    \begin{algorithmic}[1]
	\REQUIRE well-pretrained ancestry model $F_{anc}^L$.
	\ENSURE the learngene pool.
	\STATE Given $F_{anc}^L$ with output dimension $d$, narrow the number of blocks to get the learngene instance $F_{{aux}_{1}}^{l < L}$;
        \STATE Reduce the output dimension based on $F_{{aux}_{1}}^{l < L}$ to get $F_{{aux}_{2}}^{l < L}$ with output dimension $d^\prime < d$;
        \FORALL{$k=1$ to $2$}
        \FORALL{epoch = 1, \dots, 100}
        \STATE Distill $F_{anc}^L$ to $F_{{aux}_{k}}^{l}$ and train $F_{{aux}_{k}}^{l}$ with Eq.\ref{eq:loss};
        \ENDFOR
        \ENDFOR
        \STATE Select the blocks from $F_{{aux}_{k}}^{l}, k=1,2$ as learngene instances;
        \STATE Construct the learngene pool by learngene instances;
        \STATE Initialize the stitch layers;
        \FORALL{epoch = 1, \dots, 50}
        \STATE Randomly sample one path from the learngene pool;
        \STATE Train the sampled path by minimizing  Eq.\ref{qe:pred} and Eq.\ref{eq:cls}.
        \ENDFOR
    \end{algorithmic}  
\end{algorithm}


\textbf{Stitching Directions.} 
Following the establishment of the learngene pool, we proceed to build descendant models with variable sizes to satisfy diverse resource constraints. 
Within the learngene pool, we perform stitching operations from the smaller learngene instances to the larger ones. This operation aligns with SN-Net, demonstrating that stitching from smaller instances to larger ones yields enhanced stability and performance, as shown in Figure \ref{fig:overall} (c)(\romannumeral2).

\section{Experiments}
\subsection{Implementation Setting}

\textbf{Dataset.} We conduct all experiments on ImageNet-1K \cite{Russakovsky_Deng_Su_Krause_Satheesh_Ma_Huang_Karpathy_Khosla_Bernstein_etal._2015} dataset. ImageNet-1K is a large-scale image dataset designed for the classification task with 1,000 categories. It consists of a training set with 1.2 million images, and a validation set consisting of 50,000 images. 
During the training and testing phases, the initial images are resized to a resolution of $224 \times 224$.

\textbf{Architectures.} We adopt the DeiT-Base \cite{touvron2021training} as the ancestry model, initialized with pre-trained parameters from Timm \cite{rw2019timm}. 
Additionally, we create two auxiliary models: 6 blocks with 192 and 768 output dimensions.
The two auxiliary models then construct the learngene pool which contains 12 learngene instances. For convenience, we denote it as the learngene pool (12).
To further verify \textbf{Learngene Pool}, we also design larger auxiliary models: 9 blocks with 192 and 768 output dimensions. 
The two auxiliary models construct another learngene pool with 18 learngene instances, termed the learngene pool (18).

\textbf{Training Details.} We train the auxiliary models with 100 epochs and freeze the ancestry model during distillation. 
We employ 150 epochs for training the descendant models from scratch and 50 epochs to finetune the learngene pool.
The batch size is set to 128, and the initial learning rate is set to $5\times 10^{-4}$. All other hyperparameters remain consistent with the default setting of SN-Net \cite{pan2023stitchable}.

\subsection{Main Results and Analysis}

\textbf{Learngene Pool vs. SN-Net.}  We first conduct the comparison between the proposed \textbf{Learngene Pool} and SN-Net \cite{pan2023stitchable}. 
Since SN-Net stitches from the large anchors while \textbf{Learngene Pool} employs smaller instances, it is hard to establish a one-to-one correspondence between the models constructed from these approaches.
Therefore, we compare the performance of models built from the learngene pool and SN-Net under conditions of equivalent storage resource costs. 
Note that, we use the official code in \url{https://github.com/ziplab/SN-Net} to implement the experiments of SN-Net.
The comparison results are presented in Table \ref{tab:ex_compare_snnet_ours12} for a resource cost of 47.83M and in Table \ref{tab:ex_compare_snnet_ours18} for a resource cost of 70.87M. 
As demonstrated, \textbf{Learngene Pool} achieves superior performance in nearly all constructed descendant models compared to SN-Net, under both resource cost scenarios. 

\begin{table}[t]
\resizebox{\linewidth}{!}{
\begin{tabular}{c|c|c|c|c|c}
\toprule[1pt]
\multirow{2}{*}[-1ex]{Low}    & \multirow{2}{*}[-1ex]{High}  & \multicolumn{2}{c|}{Built Models}                                 & \multirow{2}{*}[-1ex]{SN-Net}             & \multirow{2}{*}[-0.5ex]{\begin{tabular}[c]{@{}c@{}}Learngene\\ Pool (12)\end{tabular}} \\ \cmidrule{3-4}
                        &                        & \multicolumn{1}{c|}{FLOPs (G)} & \multicolumn{1}{c|}{Params (M)} &                                     &                                                                                \\ \midrule
\multicolumn{1}{c|}{6} & \multicolumn{1}{c|}{0} & \multicolumn{1}{c|}{0.64}      & \multicolumn{1}{c|}{3.05}       & \multicolumn{1}{c|}{54.37}          & \multicolumn{1}{c}{\textbf{57.00}}                                            \\ \midrule
\multicolumn{1}{c|}{5} & \multicolumn{1}{c|}{1} & \multicolumn{1}{c|}{2.03}      & \multicolumn{1}{c|}{10.38}      & \multicolumn{1}{c|}{\textbf{65.66}} & \multicolumn{1}{c}{63.77}                                                     \\ \midrule
\multicolumn{1}{c|}{4} & \multicolumn{1}{c|}{2} & \multicolumn{1}{c|}{3.38}      & \multicolumn{1}{c|}{17.02}      & \multicolumn{1}{c|}{69.47}          & \multicolumn{1}{c}{\textbf{70.00}}                                            \\ \midrule
\multicolumn{1}{c|}{3} & \multicolumn{1}{c|}{3} & \multicolumn{1}{c|}{4.73}      & \multicolumn{1}{c|}{23.66}      & \multicolumn{1}{c|}{70.33}          & \multicolumn{1}{c}{\textbf{72.78}}                                            \\ \midrule
\multicolumn{1}{c|}{2} & \multicolumn{1}{c|}{4} & \multicolumn{1}{c|}{6.09}      & \multicolumn{1}{c|}{30.31}      & \multicolumn{1}{c|}{71.01}          & \multicolumn{1}{c}{\textbf{74.21}}                                            \\ \midrule
\multicolumn{1}{c|}{1} & \multicolumn{1}{c|}{5} & \multicolumn{1}{c|}{7.44}      & \multicolumn{1}{c|}{36.95}      & \multicolumn{1}{c|}{70.72}          & \multicolumn{1}{c}{\textbf{74.78}}                                            \\ \midrule
\multicolumn{1}{c|}{0} & \multicolumn{1}{c|}{6} & \multicolumn{1}{c|}{8.85}      & \multicolumn{1}{c|}{44.04}      & \multicolumn{1}{c|}{67.98}          & \multicolumn{1}{c}{\textbf{75.05}}                                            \\ \bottomrule[1pt]
\end{tabular} }
\caption{The accuracy of the built models constructed by SN-Net and Learngenen Pool with 12 instances (Learngene Pool (12)). We denote the `Low(High)' as the number of instances with low(high) output dimensions.}
\centering 
\label{tab:ex_compare_snnet_ours12}
\end{table}

\begin{table}[]
\resizebox{\linewidth}{!}{
\begin{tabular}{c|c|cc|c|c}
\toprule[1pt]
\multirow{2}{*}[-1ex]{Low} & \multirow{2}{*}[-1ex]{High} & \multicolumn{2}{c|}{Built Models}           & \multirow{2}{*}[-1ex]{SN-Net} & \multirow{2}{*}[-0.5ex]{\begin{tabular}[c]{@{}c@{}}Learngene\\ Pool (18)\end{tabular}} \\ \cmidrule{3-4}
                     &                       & \multicolumn{1}{c|}{Flops (G)} & Params (M) &                         &                                                                                \\ \midrule
9                    & 0                     & \multicolumn{1}{c|}{0.95}      & 4.38       & 56.17                   & \textbf{61.63}                                                                 \\ \midrule
8                    & 1                     & \multicolumn{1}{c|}{2.33}      & 11.71      & 65.85          & \textbf{66.95}                                                                          \\ \midrule
7                    & 2                     & \multicolumn{1}{c|}{3.69}      & 18.36      & 68.84                   & \textbf{71.69}                                                                 \\ \midrule
6                    & 3                     & \multicolumn{1}{c|}{5.04}      & 25.00      & 70.07                   & \textbf{74.38}                                                                 \\ \midrule
5                    & 4                     & \multicolumn{1}{c|}{6.39}      & 31.64      & 70.87                   & \textbf{75.69}                                                                 \\ \midrule
4                    & 5                     & \multicolumn{1}{c|}{7.75}      & 38.29      & 71.46                   & \textbf{76.42}                                                                 \\ \midrule
3                    & 6                     & \multicolumn{1}{c|}{9.10}      & 44.93      & 71.93                   & \textbf{76.95}                                                                 \\ \midrule
2                    & 7                     & \multicolumn{1}{c|}{10.45}     & 51.57      & 72.20                    & \textbf{77.13}                                                                          \\ \midrule
1                    & 8                     & \multicolumn{1}{c|}{11.80}     & 58.21      & 71.85                   & \textbf{77.32}                                                                          \\ \midrule
0                    & 9                     & \multicolumn{1}{c|}{13.26}     & 65.30      & 69.11                   & \textbf{77.42}                                                                          \\ \bottomrule[1pt]
\end{tabular}}
\caption{The accuracy of the built models constructed by SN-Net and Learngenen Pool with 18 instances (Learngene Pool (18)).}
\centering 
\label{tab:ex_compare_snnet_ours18}
\end{table}

Additionally, compared to the storage resource costs (118.4M) of SN-Net reported in \cite{pan2023stitchable}, the learngene pool with 12 instances saves around 59.6\% storage resources (118.4M vs. 47.83M), and the
learngene pool with 18 instances reduces around 40.1\% storage resources (118.4M vs. 70.87M). 
Moreover, the DeiT-based SN-Net fails to build models with parameters below 5.7M, as reported in  \cite{pan2023stitchable}. In contrast, \textbf{Learngene Pool} can construct models with 3.05M parameters from the learngene pool with 12 instances, as well as models with 4.38M parameters from the learngene pool with 18 instances, as demonstrated in the first row of Table \ref{tab:ex_compare_snnet_ours12} and Table \ref{tab:ex_compare_snnet_ours18}.

Compared to training from scratch, the learngene pool with 12 instances reduces around 6.75× training costs (150+200+50 epochs vs. 18×150 epochs), and the learngene pool with 18 instances reduces around 10.13× training costs (150+200+50 epochs vs. 27×150 epochs). Note that we train the models from scratch with 150 epochs, and the saving cost can be further enlarged when training from scratch with more epochs.

\textbf{12 vs. 18 instances Learngene Pools.} Furthermore, we compare the performance of the descendant models built from the learngene pool with 12 and 18 instances. Noteworthy, since the sizes of descendant models built from these two learngene pools are not in one-to-one correspondence, we compare the performance of descendant models within a certain parameter size range. Within each range, we select the highest-performing descendant model. Therefore, the results, as shown in Table \ref{tab:comp_lp}, are different from Table \ref{tab:ex_compare_snnet_ours12} and Table \ref{tab:ex_compare_snnet_ours18}.
We find that the descendant models tend to perform better when built from the learngene pool (18), which contains more instances in Table \ref{tab:comp_lp}. 

\subsection{Ablation Studies}
In this section, we ablate the number of blocks to calculate distillation loss functions when distilling the ancestry model to the auxiliary models and ways of initializing the stitch layers for finetuning the learngene pool. 
The training strategy is introduced in the section ``Implementation Setting.".

\textbf{The number of blocks to Distill.}
To study the effect of the number of blocks for distilling auxiliary models from the ancestry model, we consider 3 cases: 
1) without the distillation, i.e., training from scratch. 
2) only distilling the information from the last block in the ancestry model. 
3) distilling the information of three blocks in the ancestry model, as introduced in the section ``Constructing the Learngene Pool''. 
The results are listed in Table \ref{tab:abla_distill}. 

\begin{table}[]
\resizebox{\linewidth}{!}{
\begin{tabular}{c|cc|cc}
\toprule[1pt]
Built   Models & \multicolumn{2}{c|}{Learngene Pool (12)}                                 & \multicolumn{2}{c}{Learngene Pool (18)}               \\ \midrule
Params (M)      & \multicolumn{1}{c|}{Params (M)}             & Acc (\%)                   & \multicolumn{1}{c|}{Params (M)}             & Acc (\%) \\ \midrule
\textless{}= 5 & \multicolumn{1}{c|}{\multirow{7}{*}[-3ex]{47.83}} & 57.00                      & \multicolumn{1}{c|}{\multirow{7}{*}[-3ex]{70.87}} & \textbf{61.63}    \\ \cmidrule{1-1} \cmidrule{3-3} \cmidrule{5-5} 
5--15          & \multicolumn{1}{c|}{}                       & 64.30                      & \multicolumn{1}{c|}{}                       & \textbf{67.39}    \\ \cmidrule{1-1} \cmidrule{3-3} \cmidrule{5-5} 
15--25         & \multicolumn{1}{c|}{}                       & 73.42                     & \multicolumn{1}{c|}{}                       & \textbf{74.38}    \\ \cmidrule{1-1} \cmidrule{3-3} \cmidrule{5-5} 
25-35          & \multicolumn{1}{c|}{}                       & 74.54                      & \multicolumn{1}{c|}{}                       & \textbf{76.01}   \\ \cmidrule{1-1} \cmidrule{3-3} \cmidrule{5-5} 
35-45          & \multicolumn{1}{c|}{}                       & 75.19                      & \multicolumn{1}{c|}{}                       & \textbf{76.95}    \\ \cmidrule{1-1} \cmidrule{3-3} \cmidrule{5-5} 
45-55          & \multicolumn{1}{c|}{}                       & \multirow{2}{*}{Uncover} & \multicolumn{1}{c|}{}                       & \textbf{77.35}    \\ \cline{1-1} \cmidrule{5-5} 
55             & \multicolumn{1}{c|}{}                       &                            & \multicolumn{1}{c|}{}                       & \textbf{77.47}    \\ \bottomrule[1pt]
\end{tabular}}
\caption{The accuracy of the descendant models built from Learngene Pool (12) and Learngene Pool (18). `Uncover' means the target models cannot be built.}
\centering 
\label{tab:comp_lp}
\end{table}

\begin{table}[]
\resizebox{\linewidth}{!}{
\begin{tabular}{c|cc|cc}
\toprule[1pt]
\multirow{2}{*}{Number} & \multicolumn{2}{c|}{Unmatching}            & \multicolumn{2}{c}{Matching}          \\ \cmidrule{2-5} 
                        & \multicolumn{1}{c|}{6 blocks} & 9 blocks & \multicolumn{1}{c|}{6 blocks} & 9 blocks \\ \midrule
0                       & \multicolumn{1}{c|}{53.49}    & 58.96    & \multicolumn{1}{c|}{67.20}     & 68.46    \\ \midrule
1                       & \multicolumn{1}{c|}{44.27}    & 54.98    & \multicolumn{1}{c|}{\textbf{78.83}}    & \textbf{80.28}    \\ \midrule
3                       & \multicolumn{1}{c|}{\textbf{53.82}}    & \textbf{60.44}    & \multicolumn{1}{c|}{70.57}    & 75.38    \\ \bottomrule[1pt]
\end{tabular}}
\caption{The results of training the auxiliary models with various numbers of blocks used in the distillation losses. `Matching' indicates that the output dimension of the auxiliary models matches that of the ancestry model, while 'Unmatching' means a difference. `6(9) blocks' refers to auxiliary models with 6(9) blocks.}
\centering 
\label{tab:abla_distill}
\end{table}

It can be found that for auxiliary models with lower output dimensions than the ancestry model, the performance of the auxiliary models can be enhanced when incorporating three blocks to distill, as shown in the column `Unmatching' in Table \ref{tab:abla_distill}. 
We speculate that the difference in output dimensions results in the loss of critical information during distillation. Therefore, more blocks are required to fully distill information to the auxiliary model.
Moreover, the performance of the auxiliary model is even lower than training from scratch when taking one block to distill. 
This implies that taking one block for distillation introduces a considerable amount of noise from the high-dimensional space to the low-dimensional space, resulting in a decline in the accuracy of the auxiliary models.

Conversely, for auxiliary models with identical output dimensions to the ancestry model, the distillation of individual blocks can enhance the learning of the auxiliary model, as indicated in the column `Matching' in Table \ref{tab:abla_distill}. 
This can be attributed to the fact that the same output dimension space facilitates the accurate distillation of critical information from the ancestry model to the auxiliary models, while more blocks introduce more noise.

\begin{figure}[t]
    \centering
    \includegraphics[width=0.45\textwidth]{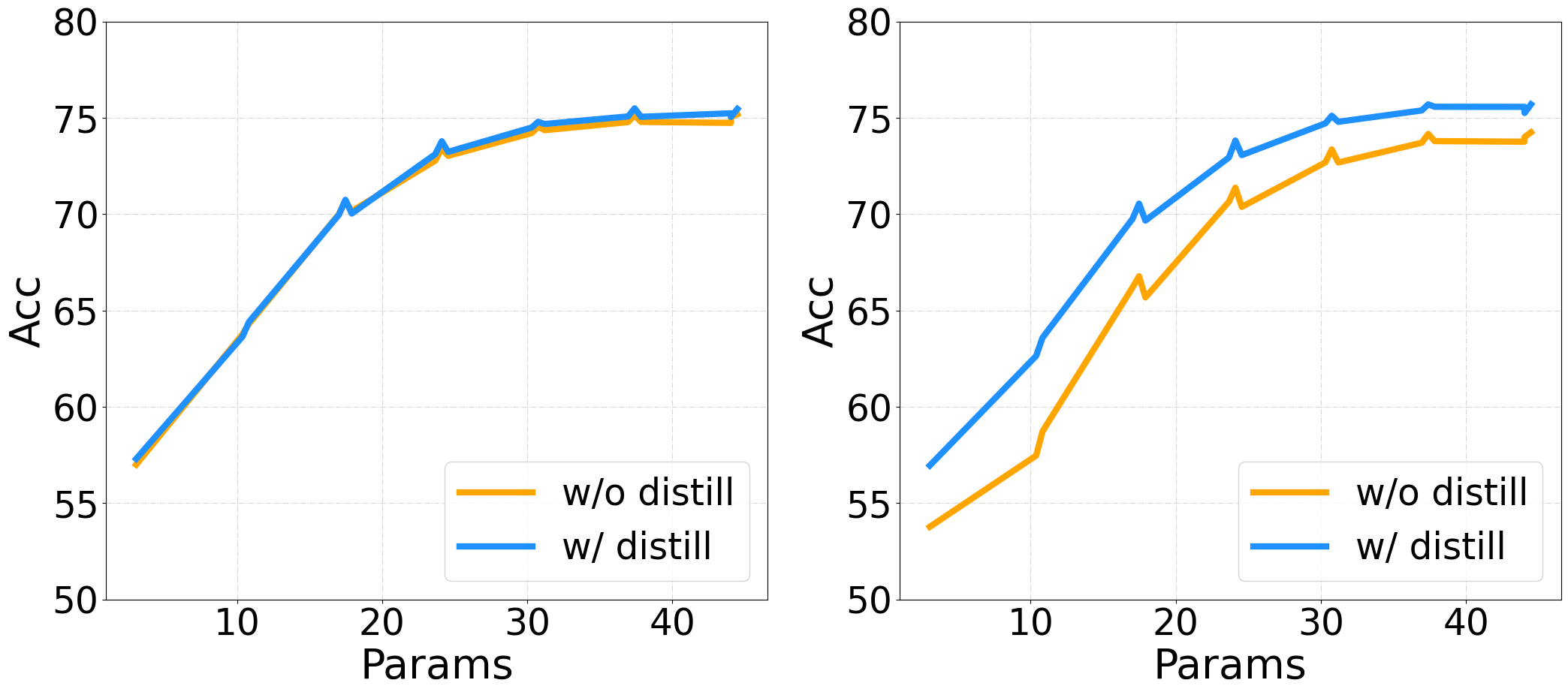}
    \caption[Figure]{The performance of the descendant models built from the learngene pool which is fine-tuned with distillation and without distillation. Left: the stitch layers are initialized by the block-based transformation matrices. Right: the stitch layers are initialized by the least square method.}
    \label{fig:distill}
\end{figure}

\textbf{Fine-tuning the Learngene Pool with or without distillation.}
To validate the application of distillation during fine-tuning the learngene pool, we compare the impact of distillation on the performance of the built descendant models, as depicted in Figure \ref{fig:distill}. We find that when the stitch layers are initialized with our block-based transformation matrices method, there is only marginal enhancement in descendant model performance, as shown in Figure \ref{fig:distill} (Left). 
Initializing the stitch layers by the least square method results in a significant performance boost for the descendant models, as shown in Figure \ref{fig:distill} (Right).
This indicates that for our block-based transformation matrices method, it is unnecessary to perform distillation when fine-tuning the learngene pool.
However, for the least squares method, distillation remains a crucial step during fine-tuning the learngene pool.

\begin{table}[]
\resizebox{\linewidth}{!}{
\begin{tabular}{cccc}
\toprule[1pt]
\multicolumn{2}{c|}{Built Models}                                & \multicolumn{2}{c}{Initialization Way}     \\ \midrule
 \multicolumn{1}{c|}{FLOPs (G)} & \multicolumn{1}{c|}{Params (M)} & \multicolumn{1}{c|}{LS}    & TM           \\ \midrule
\multicolumn{4}{c}{Learngene Pool (12)}                                                                       \\ \midrule
\multicolumn{1}{c|}{0.64}      & \multicolumn{1}{c|}{3.05}       & \multicolumn{1}{c|}{56.95} & \textbf{57.00} \\ \midrule
\multicolumn{1}{c|}{2.03} & \multicolumn{1}{c|}{10.38}  & \multicolumn{1}{c|}{62.64} & \textbf{63.77} \\ \midrule
\multicolumn{1}{c|}{2.13}      & \multicolumn{1}{c|}{10.82}      &  \multicolumn{1}{c|}{63.59} & \textbf{64.30} \\ \midrule
\multicolumn{1}{c|}{3.38}      & \multicolumn{1}{c|}{17.02}      &  \multicolumn{1}{c|}{69.76} & \textbf{70.00} \\ \midrule
\multicolumn{1}{c|}{3.48}      & \multicolumn{1}{c|}{17.47}      &  \multicolumn{1}{c|}{70.55} & \textbf{70.70} \\ \midrule
\multicolumn{4}{c}{Learngene Pool (18)}                                                                       \\ \midrule
 \multicolumn{1}{c|}{0.95}      & \multicolumn{1}{c|}{4.38}       &\multicolumn{1}{c|}{60.73} & \textbf{61.63} \\ \midrule
 \multicolumn{1}{c|}{2.33}      &\multicolumn{1}{c|}{11.71}      & \multicolumn{1}{c|}{65.07} & \textbf{66.95} \\ \midrule
 \multicolumn{1}{c|}{2.44}      &\multicolumn{1}{c|}{12.16}      & \multicolumn{1}{c|}{65.99} & \textbf{67.39} \\ \midrule
\multicolumn{1}{c|}{3.79}      & \multicolumn{1}{c|}{18.80}      &  \multicolumn{1}{c|}{71.63} & \textbf{72.20} \\ \midrule
 \multicolumn{1}{c|}{5.04}      & \multicolumn{1}{c|}{25.00}      & \multicolumn{1}{c|}{73.81} & \textbf{74.38} \\
\bottomrule[1pt]
\end{tabular}}
\caption{Accuracy comparison of descendant models built from the learngene pool with different stitch layer initialization methods: the least square methods (LS) and our block-based transformation matrices (TM).}
\centering 
\label{tab:ex_init_way}
\end{table}

\textbf{Initialization Ways of the Stitch Layers.}
We ablate the initialization ways for stitch layers in the learngene pool: the least squares method (LS) in SN-Net, and our block-based transformation matrices (TM). 
As shown in Figure \ref{fig:distill} (left), the learngene pool with TM-initialized stitch layers constructs competitive descendant models even without distillation during fine-tuning. 
Conversely, the LS initialization results in building descendant models with inferior performances before distillation, as shown in Figure \ref{fig:distill} (right).  
Furthermore, We also compare the accuracy of descendant models built from the learngene pool where the stitch layers are initialized with LS and TM methods in Table \ref{tab:ex_init_way}. 
As it indicates, taking TM to initialize the stitch layers in the learngene pool leads to the construction of superior descendant models. 
These verify the effectiveness of TM for initializing stitch layers in this study.

\section{Conclusion}
In this paper, we propose a novel method called \textbf{Learngene pool} to build variable-sized descendant models for various resource constraints by inserting stitch layers among learngene instances in the learngene pool. 
To achieve this, we distill the larger model into multiple smaller auxiliary models. 
In this way, the auxiliary models can extract critical knowledge from the larger model. 
Then, we select all blocks in the auxiliary models as learngene instances to construct the learngene pool. The stitch layers in it are initialized by the block-based transformation matrices during the training of the auxiliary models.
Since the learngene pool consists of learngene instances with varying output dimensions, stitching them results in variable-sized descendant models. 
Compare to SN-Net, the learngene pool consumes fewer storage resources. 
Moreover, the smaller learngene instances enable to build the smaller descendant models, which adapt to low resource constraints. 
Finally, the proposed way of initializing the stitch layers enables the learngene pool to build descendant models with superior performances.
 
{\small
\bibliography{aaai24}
}

\end{document}